\documentclass[conference]{IEEEtran}
\IEEEoverridecommandlockouts
\usepackage{amsmath,amssymb,amsfonts}
\usepackage{algorithmic}
\usepackage{graphicx}
\usepackage{textcomp}
\usepackage{authblk}
\usepackage{xcolor}
\usepackage{threeparttable}
\usepackage{dsfont}
\def\BibTeX{{\rm B\kern-.05em{\sc i\kern-.025em b}\kern-.08em
    T\kern-.1667em\lower.7ex\hbox{E}\kern-.125emX}}
\usepackage[
backend=biber,
style=ieee,
]{biblatex}
\begin{document}

\title{Spiral Contrastive Learning: An Efficient 3D Representation Learning Method for Unannotated CT Lesions
\thanks{This work was supported by National Natural Science Foundation of China (62106248), and Ningbo Natural Science Foundation (202003N4270).}
\thanks{*Corresponding author: Jinpeng Li (lijinpeng@ucas.ac.cn).}
}

\author[1, 2]{\textbf{Penghua Zhai}}
\author[2]{\textbf{Enwei Zhu}}
\author[2]{\textbf{Baolian Qi}}
\author[2]{\textbf{Xin Wei}}
\author[1, 2]{\textbf{Jinpeng Li}*}
\affil[1]{HwaMei Hospital, University of Chinese Academy of Sciences (UCAS), Ningbo, China}%
\affil[2]{Ningbo Institute of Life and Health Industry, UCAS, Ningbo, China}%

\maketitle

\begin{abstract}
Computed tomography (CT) samples with pathological annotations are difficult to obtain. As a result, the computer-aided diagnosis (CAD) algorithms are trained on small datasets (e.g., LIDC-IDRI with 1,018 samples), limiting their accuracies and reliability. In the past five years, several works have tailored for unsupervised representations of CT lesions via two-dimensional (2D) and three-dimensional (3D) self-supervised learning (SSL) algorithms. The 2D algorithms have difficulty capturing 3D information, and existing 3D algorithms are computationally heavy. Light-weight 3D SSL remains the boundary to explore. In this paper, we propose the spiral contrastive learning (SCL), which yields 3D representations in a computationally efficient manner. SCL first transforms 3D lesions to the 2D plane using an information-preserving spiral transformation, and then learn transformation-invariant features using 2D contrastive learning. For the augmentation, we consider natural image augmentations and medical image augmentations. We evaluate SCL by training a classification head upon the embedding layer. Experimental results show that SCL achieves state-of-the-art accuracy on LIDC-IDRI (89.72\%), LNDb (82.09\%) and TianChi (90.16\%) for unsupervised representation learning. With 10\% annotated data for fine-tune, the performance of SCL is comparable to that of supervised learning algorithms (85.75\% vs. 85.03\% on LIDC-IDRI, 78.20\% vs. 73.44\% on LNDb and 87.85\% vs. 83.34\% on TianChi, respectively). Meanwhile, SCL reduces the computational effort by 66.98\% compared to other 3D SSL algorithms, demonstrating the efficiency of the proposed method in unsupervised pre-training.
\end{abstract}


\section{Introduction}
With the renaissance of deep learning, computer-aided diagnosis (CAD) systems for computed tomography (CT) have achieved many successful applications \cite{yanase2019systematic, zhang2019computer, chan2020computer}. However, Existing CAD systems are mostly developed based on supervised learning which heavily rely on lesion-level annotations. Meanwhile, these annotations are often scarce because only expert experienced radiologists can annotate the lesions. This drives our interest to the unsupervised representation learning. Recent studies show that self-supervised learning (SSL) is an effective approach for learning representations. In the past five years, SSL methods are mostly tailored for unsupervised representations via two-dimensional (2D) images \cite{noroozi2016unsupervised, chen2020simple, he2020momentum, zhai2021mvcnet}. However, it is difficulty for 2D SSL algorithms to capture three-dimensional (3D) information of CT scans. Therefore, several 3D SSL are explored to learn the 3D representations of lesions \cite{zhou2019models, zhu2020rubik}. \textit{Nevertheless, existing 3D SSL algorithms suffer from the computationally heavy and slow convergence problems}.

To address these issues, we propose a novel light-weight spiral contrastive learning (SCL) method to efficiently learn 3D representations for unannotated CT lesions with fewer computational resources and parameters. Unlike existing 3D SSL methods, we feed a transformed 2D view into the network, which is converted from a 3D lesion volume by the spiral transformation \cite{chen2020combined}. To fully learn the 3D representations and spatial information, the spiral transformation has the following property: The 2D views should preserve the 3D features of lesions as much as possible. Therefore, the proposed SCL is able to learn the 3D representations by the spiral transformation. We regard the views of the same lesion as positive pairs, and the views of different lesions as negative pairs. We learn representations by minimizing the contrastive loss to obtain a embedding space with the property of within-lesion compactness and between-lesion separability. We validate the effectiveness of our method on three lung CT datasets. The experimental results show that the proposed SCL outperforms state-of-the-art SSL methods.

The contributions of this paper are as follows: 1) An information-preserving spiral transformation is introduced to convert each 3D lesion to a 2D view. The 2D view can preserve the correlation between 3D lesion adjacent pixels and retain the spatial relationship of texture features for the 3D volume. 2) We propose a novel spiral contrastive learning (SCL) method, which yields 3D transformation-invariant representations in a computionally efficient manner. It is a light-weight 3D SSL algorithm. 3) We evaluate the proposed SCL by training a simple classification head upon the embedding layer. Experimental results show that SCL achieves state-of-the-art accuracy on LIDC-IDRI, LNDb and TianChi for unsupervised representation learning. With 10\% annotated data for fine-tuning, the performance of SCL is comparable to that of supervised learning algorithms. Meanwhile, SCL reduces the computational effort by 66.98\% compared to other 3D SSL algoithms, demonstrating the efficiency of the proposed method in unsupervised pre-training.

\begin{figure*}[t]
    \centering
    \includegraphics[width=0.9\textwidth]{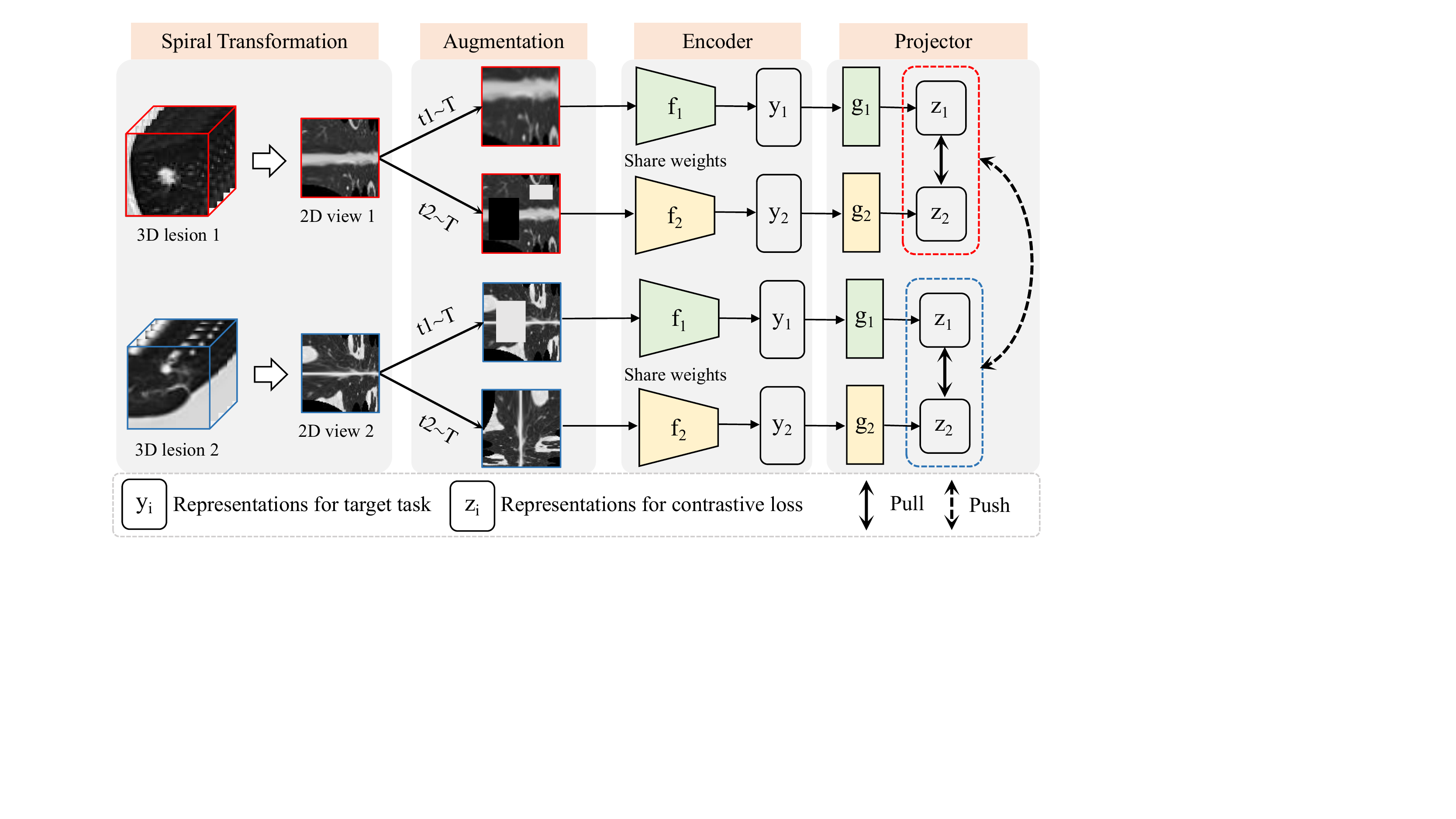}
    \caption{Illustration of the training stage of the SCL. Each lesion volume is converted to a 2D view by the information-preserving spiral transformation, and two augmented views are generated by the augmentation $ \boldsymbol{t}_{i} $, where $ i = 1, 2 $. Then, we construct a light-weight contrastive learning to generate transformation-invariant representations for each augmented view. We make views of the same lesions attract each other and views of different lesions repeal each other to learn discriminative features. Being optimized by a contrastive loss, the shared embedding space is endowed with well local-aggregating properties and the spread-out properties of representations are preserved. $ \boldsymbol{y}_{i} $ and $ \boldsymbol{z}_{i} $represent the features learned from encoder $ f_{i} $ and $ g_{i} $}
    \label{scl_train}
\end{figure*}

\section{Related Works}

\subsection{Self-supervised Learning}
SSL, especially contrastive learning, is an extensively-studied area in recent years due to its effectiveness in unsupervised representation learning \cite{jing2020self, liu2021self}. Gidaris et al. \cite{komodakis2018unsupervised} learned image features by recognizing the rotation of the input image, and they proved that this simple task actually provides a very powerful supervision signal for semantic feature learning. He et al. \cite{he2020momentum} facilitated contrastive learning by building a dynamic dictionary with a queue and a moving-averaged encoder. Chen et al. \cite{chen2020simple} proposed a simple contrastive visual representation learning framework by aggregating the views of the same image and separating the other views. Furthermore, a series of transformation-based methods \cite{tian2020contrastive, chen2020improved} and pretext task \cite{noroozi2016unsupervised, zhang2016colorful} are proposed to achieve SSL.

SSL methods are also used to achieve medical image representation learning \cite{zhou2019models, li2020self, li2021multi}. Chen et al. \cite{chen2019self} proposed a novel context restoration method for fetal 2D ultrasound detection to learn semantic image features. However, the 2D network has difficulty in capturing 3D information of lesions. Zhou et al. \cite{zhou2019models} learned the common anatomical representation automatically by using the sophisticated yet recurrent anatomy in medical images as supervision signals. Zhu et al. \cite{zhu2020rubik} proposed a pretext task, i.e., Rubik’s cube+, to force networks to learn translation and rotation invariant features from the 3D medical data. Li et al. \cite{li2020self} presented a novel multi-modal self-supervised feature learning method by capturing the semantically shared information across different modalities and the apparent visual similarity between patients for retinal disease diagnosis. However, existing 3D SSL methods are computationally heavy, such as the number of parameters of Rubik's cube+ \cite{zhu2020rubik} is $ 84M $, while the parameters of SimCLR \cite{chen2020simple} and MoCo \cite{he2020momentum} are less than $ 12M $. 2D contrastive learning has fewer 66.98\% parameters than 3D contrastive learning with same backbone (such as ResNet18 and 3D ResNet18 \cite{he2016deep}), the 2D. It is necessary to further explore the SSL by combining the advantages of 2D and 3D networks to learn spatial information of CT scans.

\subsection{Automatic Lung CT Diagnostic Models}
CAD systems are used to assist radiologists in interpreting disease-related information in medical images \cite{yanase2019systematic, zhang2019computer}. Considering the 3D characteristics of lesions, Setio et al. \cite{setio2016pulmonary} achieved the lung nodule diagnosis by composing the 3D lesion volume into nine views. Although the method achieved good results by fusing multiple 2D views, the multi-view method is not always suitable for radiologists. In the process of CT screening, radiologists usually observe the complete 3D lesions in CT scans for diagnosis. Therefore, Mei et al. \cite{mei2021sanet} proposed a 3D slice-aware network (SANet) for pulmonary nodule detection and introduced a multi-scale feature maps to reduce the false positive. Shen et al. \cite{shen2019interpretable} presented a novel interpretable deep hierarchical semantic CNN by combining low-level and high-level features to predict malignant nodules. However, compared to models trained on natural images, the medical models are trained on the small datasets (e.g. LIDC-IDRI \cite{armato2011lung}) with limited accuracy and generalization capabilities \cite{xie2018knowledge}. It is a challenge to achieve the high-quality 3D CAD systems with limited labeled samples.

\begin{figure}
    \centering
    \includegraphics[width=0.8\linewidth]{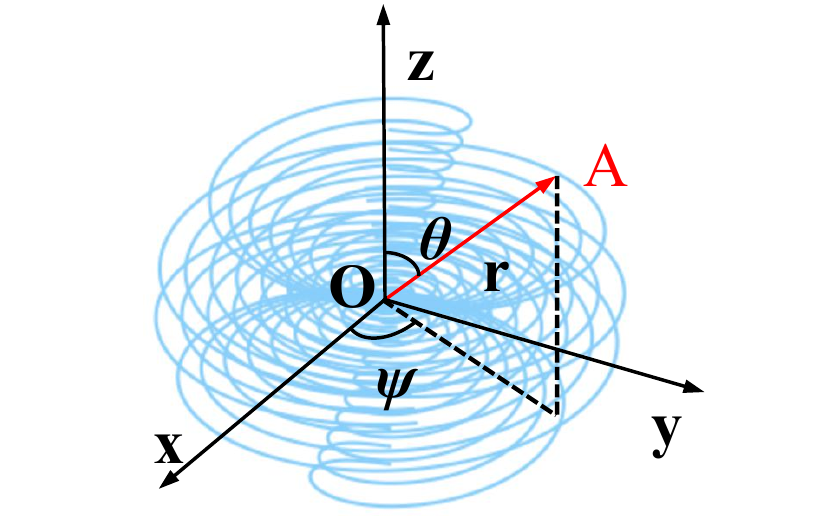}
    \caption{Coordinate system of spiral transformation. The coordinate origin $ O $ is the center-point of spiral transformation. And the spiral line is calculated by $ \Theta $, $ \Psi $ and $ r $.}
    \label{spiral_trans}
\end{figure}

\section{Methods}
In this section, we illustrate the novel light-weight spiral contrastive learning (SCL) method as shown in Fig. \ref{scl_train}. Assuming that the suspicious lesions have been detected, and we first convert each 3D lesion volume to a 2D view by the spiral transformation. We then learn 3D transformation-invariant representations with a computationally efficient manner. We finally evaluate the representations by training a simple classification head.

\subsection{Information-preserving Spiral Transformation}
Typically, 3D networks are better than 2D networks for learning spatial information of lesions, while 3D networks are computationally heavy. Inspired by the SpiralTransform \cite{chen2020combined}, we introduce a information-preserving spiral transformation to convert each 3D lesion to a 2D view, as shown in Fig. \ref{spiral_trans}. Therefore, a light-weight contrastive learning method can be developed to obtain the 3D representations. During the transformation, the method preserves the correlation between original adjacent pixels. Therefore, the 2D view can retain the spatial relationship of texture features from the 3D lesion volume.


We denote the dataset by $ S = \{(X_i, y_i)\}_{i=1}^{N} $, where $ X_i \in \mathbb{R}^{d \times w \times h} $ is the $ i $-th 3D lesion volume, $ y_i $ denotes the label of $ X_i $, and the $ N $ is the number of samples. We select the center point of the lesion volume $ X_i \in \mathbb{R}^{d \times w \times h} $ as the midpoint $ O $ of the spiral transformation, and the space rectangular coordinate system is established with midpoint $ O $. The maximum radius $ R $ of the transformation depends on the maximum distance from the tumor edge to the $ O $ point. The spiral line \textit{A} is determined by an azimuth angle $ \Psi $, an elevation angle $ 1 - \Theta $, and the distance to $ O $ in the 3D space. The coordinates of $ A $ can be expressed as:
\begin{equation}
    \left\{\begin{array} {l}
        {x = r \operatorname {sin} \Theta \operatorname {cos} \Psi} \\
        {y = r \operatorname {sin} \Theta \operatorname {sin} \Psi \quad \text {where}} \\
        {z = r \operatorname {cos} \Theta}
        \end{array} 
    \left\{\begin{array}{l}
    0 \leq \Theta \leq \pi / 2 \\
    0 \leq \Psi \leq 2 \pi \\
    -R \leq r \leq R
    \end{array}\right.\right. ,
\end{equation}

Let the circle on the equator have $ 2 M $ sampling points, and the arc length $ d $ of the adjacent sampling points is defined as the distance between two points on the equator \cite{wang2007segmentation}. Therefore, the number of sampling points in a horizontal plane corresponding to angle $ \Theta $ is: 
\begin{equation}
    2 \pi r |sin\Theta| / d = 2 \pi r |sin\Theta| / (\pi r / M) = 2 M |sin\Theta| ,
\end{equation}
where the $ \Theta $ is divided into $ M $ angles evenly within the value range. In the case of a specified radius, if $ M $ is large enough, the total number of sampling points on a spiral line can be calculated as:
\begin{equation}
\begin{aligned}
m &= \int_{0}^{M} 2 M \sin \frac{k \pi}{2 M} d k = 2 M \int_{0}^{M} \frac{2 M}{\pi} \sin \frac{k \pi}{2 M} d \frac{k \pi}{2 M} \\
&= \frac{4 M^{2}}{\pi} \int_{0}^{\pi / 2} \sin x d x = \frac{4 M^{2}}{\pi} ,
\label{equ_m}
\end{aligned}
\end{equation}

The size of the transformed 2D view is $ 2 R \times m $ due to the number of sampling points is same for each $ r $ from \eqref{equ_m}. We set the $ R $ to 46, and the $ M $ to 12 in this paper. The gray value of the sampling points is calculated by trilinear interpolation, whose coordinates in the 3D space are then mapped to the original matrix position.

\begin{figure}[t]
    \centering
    \includegraphics[width=\linewidth]{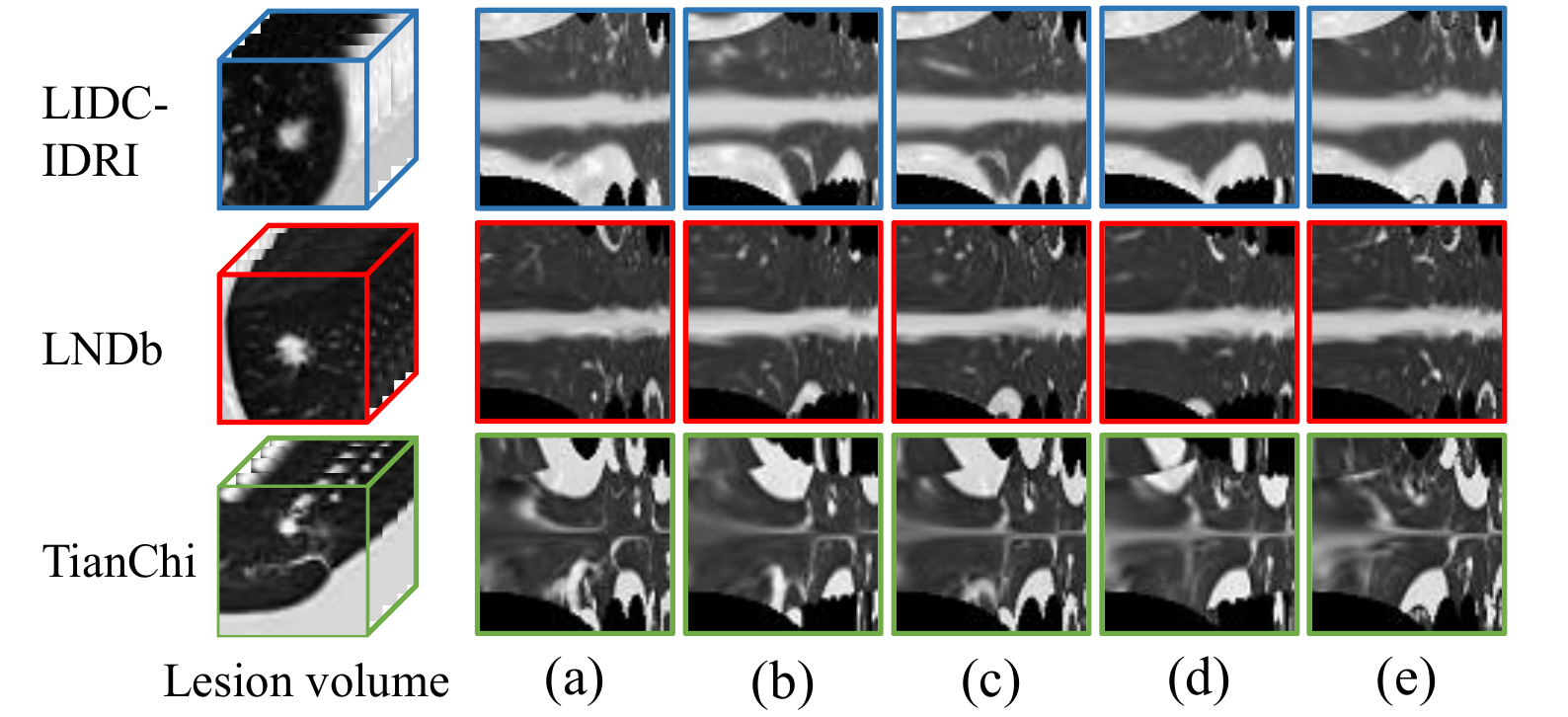}
    \caption{The views generated by the spiral transformation in different coordinate systems for each dataset. For each column, the transformed 2D views are generated from the same 3D lesion. The rows from (b) to (e) show that the coordinate system is rotated by 50, 90, 140 and 180 degrees relative to (a) in the x-y plane, respectively.}
    \label{augmented_views}
\end{figure}

The result of spiral transformation depends on the spatial rectangular coordinate system and the parameters of the transformation \cite{chen2020combined}. For the same 3D data, once the direction of the coordinate system in the spiral transformation is modified, a new transformed image can be obtained. Fig. \ref{augmented_views} simulates the data generation results using the spiral transformation in different directions.

\subsection{Light-weight SSL Representation Learning}
Contrastive learning aims to construct a latent embedding space that separates samples from different clusters in an unsupervised manner. The 2D algorithms have difficulty capturing 3D information, and existing 3D algorithms are computationally heavy. Therefore, we design a light-weight contrastive learning to achieve 3D lesion representation learning. In this section, we describe the process of improving the similarity within lesions and separability between lesions by a contrastive loss. As shown in Fig. \ref{scl_train}. The two networks have the same structure and share weights. Encoder $ f(\cdot) $ is used to learn embedding space, and projector $ g(\cdot) $ is used to eliminate the semantically irrelevant low-level information from the representation.

Given an input volume $ X $, a 2D view $ \boldsymbol{x} $ is generated by the spiral transformation. Then, two views $ \boldsymbol{v}_{1} = \boldsymbol{t}{1}(\boldsymbol{x}) $ and $ \boldsymbol{v}_{2} = \boldsymbol{t}_{2}(\boldsymbol{x}) $ are generated by two different sets of augmentations $ \boldsymbol{t}{1} $ and $ \boldsymbol{t}{2} $, and the two set of augmentations are randomly sampled from the distribution $ \boldsymbol{T} $. The two transformed 2D views are treated as the input images for the network. Specifically, the distribution $ \boldsymbol{T} $ is divided into two categories: medical image augmentation (MIA) and natural image augmentation (NIA). The NIA combines with standard data augmentation methods such as cropping, resizing, flipping, and Gaussian blur \cite{chen2020simple}. The MIA includes in-painting, out-painting, local pixel shuffling and non-linear transformation \cite{zhou2019models}.

The aim of the encoder $ f(\cdot) $ is to extract representation vectors from augmented samples. The representation $ \boldsymbol{y}_{i} = f(\boldsymbol{x}_i) $, where $ \boldsymbol{y}_{i} \in \mathbb{R}^{d} $ obtained from the output of the average pooling layer. Then, a projection head $ g(\cdot) $ maps the representation to the space where contrastive loss is applied. We use a multi-layer perceptron (MLP) with two hidden layers as the projection head to obtain $ \boldsymbol{z}_{i} = g(\boldsymbol{y}_{i}) = g(f(\boldsymbol{x}_{i})) $. We define the contrastive loss on $ \boldsymbol{z}_{i} $ rather than $ \boldsymbol{y}_{i} $. The importance of using the representation before the nonlinear projection is due to loss of information induced by the contrastive loss. The $ \boldsymbol{z}_{i} = g(\boldsymbol{y}_{i}) $ is trained to be invariant to data transformation. Thus, $ g(\cdot) $ can remove information that may be useful for the downstream task, such as the color information. More information can be maintained in $ \boldsymbol{y}_{i} $ by using the nonlinear transformation $ g(\cdot) $ \cite{chen2020simple}. The projector $ g(\cdot) $ includes a linear layer with an output size of 512, a batch normalization layer, a ReLU activation function, and a linear layer with an output size of 128.

Finally, we minimize the contrastive loss by aggregating the positive pair and separating the negative pair. Considering $ N $ samples in a batch, there are $ 2 N $ views augmented by the $ \boldsymbol{t}{1} $ and $ \boldsymbol{t}{2} $. We treat the two augmented views from the same lesion as positive pair, the other $ 2 (N - 1) $ augmented views as negative pair. We use the cosine similarity to represent the similarity of the representations of different views:
\begin{equation}
    \operatorname{sim}(\boldsymbol{p}_1, \boldsymbol{p}_1)=\boldsymbol{p}_{1}^{\top} \boldsymbol{p}_{2} /\|\boldsymbol{p}_{1}\|\|\boldsymbol{p}_{2}\|
\end{equation}
where $ \boldsymbol{p}_{1}, \boldsymbol{p}_{2} $ denote the normalized features of $ \boldsymbol{v}_{1} $ and $ \boldsymbol{v}_{2} $ from the projection head, respectively. Then, the loss function for a positive pair of examples $ (i, j) $ is defined as:
\begin{equation}
\ell_{i, j}=-\log \frac{\exp \left(\operatorname{sim}\left(\boldsymbol{z}_{i}, \boldsymbol{z}_{j}\right) / \tau\right)}{\sum_{k=1}^{2 N} \mathds{1}_{[k \neq i]} \exp \left(\operatorname{sim}\left(\boldsymbol{z}_{i}, \boldsymbol{z}_{k}\right) / \tau\right)}
\end{equation}
where $ \mathds{1}_{[k \neq i]} \in \{0, 1\} $ is an indicator function evaluating to 1 if $ k \neq i $ and $ \tau $ denotes a temperature parameter. The final loss is computed across all positive pairs in a batch, both $ (i, j) $ and $ (j, i) $.

\renewcommand{\arraystretch}{1.4}
\begin{table*}[h]
	\centering
	\setlength{\tabcolsep}{14pt}
	\begin{threeparttable}
		\caption{Comparison of SCL with state-of-the-art SSL methods on the three datasets ($ Mean(\%) \pm std (\%) $)}
		\label{table_linear_cls}
		\begin{tabular}{l c c c c c c}
			\hline\hline
			 & \multicolumn{2}{c}{LIDC-IDRI} & \multicolumn{2}{c}{LNDb} & \multicolumn{1}{c}{TianChi} \\
			\hline
			 & \textbf{AUC} & \textbf{Accuracy} & \textbf{AUC} & \textbf{Accuracy} & \textbf{Accuracy} \\
			\hline\hline
			\multicolumn{4}{l}{Natural Image Augmentation} \\
			\hline
			Context \cite{doersch2015unsupervised} & $ 64.93 \pm 1.60 $ & $ 56.88 \pm 0.97 $ & $ 64.57 \pm 0.32 $ & $ 62.50 \pm 0.58 $ & $ 64.13 \pm 1.05 $ \\
			RotNet \cite{komodakis2018unsupervised} & $ 64.96 \pm 1.14 $ & $ 56.61 \pm 2.21 $ & $ 63.80 \pm 0.59 $ & $ 58.40 \pm 0.68 $ & $ 60.22 \pm 0.39 $ \\
			MoCo \cite{he2020momentum} & $ 71.07 \pm 0.11 $ & $ 71.29 \pm 0.13 $ & $ 66.09 \pm 0.48 $ & $ 67.97 \pm 0.46 $ & $ 73.44 \pm 0.67 $ \\
			MoCo V2 \cite{chen2020improved} & $ 71.88 \pm 0.33 $ & $ 71.83 \pm 0.25 $ & $ 68.20 \pm 0.26 $ & $ 70.31 \pm 0.44 $ & $ 76.13 \pm 0.72 $ \\
			SimCLR \cite{chen2020simple} & $ 78.58 \pm 0.52 $ & $ 79.04 \pm 0.50 $ & $ 69.71 \pm 0.69 $ & $ 73.23 \pm 0.91 $ & $ 79.08 \pm 0.56 $ \\
			BYOL \cite{grill2020bootstrap} & $ 78.35 \pm 0.22 $ & $ 65.19 \pm 0.51 $ & $ 63.12 \pm 0.33 $ & $ 67.97 \pm 0.70 $ & $ 63.04 \pm 0.85 $ \\
			SimSiam \cite{chen2021exploring} & $ 76.39 \pm 0.93 $ & $ 69.11 \pm 1.12 $ & $ 68.20 \pm 0.44 $ & $ 71.88 \pm 0.46 $ & $ 73.93 \pm 0.43 $ \\
			\hline\hline
			\multicolumn{4}{l}{Medical Image Augmentation} \\
			\hline
			MoCo \cite{he2020momentum} & $ 73.27 \pm 0.55 $ & $ 74.15 \pm 0.53 $ & $ 70.44 \pm 0.15 $ & $ 70.02 \pm 0.21 $ & $ 75.63 \pm 0.34 $ \\
			MoCo V2 \cite{chen2020improved} & $ 78.45 \pm 0.85 $ & $ 78.69 \pm 0.55 $ & $ 70.30 \pm 0.23 $ & $ 70.35 \pm 0.35 $ & $ 77.13 \pm 0.65 $ \\
			SimCLR \cite{chen2020simple} & $ 79.98 \pm 0.50 $ & $ 80.31 \pm 0.71 $ & $ 71.21 \pm 0.96 $ & $ 75.00 \pm 1.47 $ & $ 80.72 \pm 0.93 $ \\
			BYOL \cite{grill2020bootstrap} & $ 70.49 \pm 0.18 $ & $ 65.77 \pm 0.94 $ & $ 69.06 \pm 0.26 $ & $ 71.87 \pm 0.31 $ & $ 69.27 \pm 0.33 $ \\
			SimSiam \cite{chen2021exploring} & $ 77.10 \pm 1.12 $ & $ 70.32 \pm 1.69 $ & $ 72.30 \pm 0.36 $ & $ 74.72 \pm 0.27 $ & $ 75.38 \pm 0.59 $ \\
            Models Genesis \cite{zhou2019models} & $ 76.29 \pm 2.33 $ & $ 65.83 \pm 1.56 $ & $ 61.73 \pm 0.34 $ & $ 61.94 \pm 0.30 $ & $ 63.93 \pm 0.43 $ \\
			Rubik’s Cube+ \cite{zhu2020rubik} & $ 82.07 \pm 0.44 $ & $ 81.21 \pm 0.16 $ & $ 75.63 \pm 0.51 $ & $ 66.67 \pm 0.39 $ & $ 76.14 \pm 0.73 $ \\
			Restoration~ \cite{chen2019self} & $ 85.60 \pm 0.31 $ & $ 78.75 \pm 0.84 $ & $ 74.09 \pm 0.51 $ & $ 67.71 \pm 0.60 $ & $ 79.27 \pm 0.95 $ \\
			\hline\hline
			Ours \\
			\hline
			\textbf{SCL} (Ours) & $ \mathbf{93.89} \pm 0.60 $ & $ \mathbf{89.72} \pm 0.38 $ & $ \mathbf{77.95} \pm 0.78 $ & $ \mathbf{82.09} \pm 0.67 $ & $ \mathbf{90.16} \pm 0.02 $ \\
			\hline\hline
		\end{tabular}
	\end{threeparttable}
\end{table*}

\subsection{3D Lesion Diagnosis Task}
After the pre-training of the SCL, we evaluate the encoder by a lesion diagnosis task, assuming that the detection of suspicious lesions has been completed. A classification head $ p(\cdot) $ is designed to achieve the task. The classification head only includes a batch normalizatin layer and a fully connected layer, and the input of $ p(\cdot) $ is the output of the encoder $ f(\cdot) $ rather than the projector $ g(\cdot) $.
We apply the SCL to lung CT dataset 10 times independently, with the 10-fold cross validation. The performance is assessed by the mean and standard deviation of accuracy and area under the receiver operator curve (AUC) \cite{zhai2020multi, xie2018knowledge}. Accuracy shows the the proposed model can correctly distinguish between malignant and benign nodules and defines as :
\begin{equation}
    Accuracy = \frac{TP + TN}{TP +TN + FP + FN} ,
\end{equation}
where $ TP $ is true positive; $ TN $ is true negative; $ FP $ is false positive; $ FN $ is false negative. AUC takes into account the sensitivity and specificity in a comprehensive manner.

\section{Experiments and Results}
In order to evaluate the proposed SCL, we conduct a series of experiments on lung CT datasets and compared SCL with state-of-the-art SSL models.

\subsection{Datasets}
\subsubsection{LIDC-IDRI}
The LIDC-IDRI \footnote{https://wiki.cancerimagingarchive.net/display/Public/LIDC-IDRI} is a commonly-used CT dataset, which contains 1,018 CT scans for lung nodule diagnosis \cite{armato2011lung}. The malignancy of each nodule is evaluated with a 5-point scale from benign to malignant by up to four experienced radiologists in two stages. In this study, scans with thickness thicker than $ 2.5 mm $ are eliminated as this could easily lead to the omission of small nodules \cite{setio2016pulmonary}. We calculate the mean malignancy ($ l $) of a nodule which is annotated by at least three radiologists, and annotate a nodule whose $ l < 3 $ as benign, a nodule whose $ l = 3 $ as uncertain and a nodule whose $ l > 3 $ as malignant. To reduce the impact of uncertain evaluation, we exclude all uncertain lung nodules from the dataset. Finally, there are totally 369 benign and 335 malignant nodules.

\subsubsection{LNDb}
The LNDb \footnote{https://lndb.grand-challenge.org/Data/} dataset contains a total of 294 CT scans \cite{pedrosa2019lndb}. Each CT scan is read by at least one radiologist with a 5-point scale from benign to malignant. We process the LNDb in the LIDC-IDRI way, with the difference that we calculate the mean malignancy ($ l $) of all annotated nodules, since few nodules annotated by at least three radiologists. Finally, there are 451 benign and 768 malignant nodules.

\subsubsection{TianChi}
The TianChi \footnote{https://tianchi.aliyun.com/competition/entrance/231724/information?lang\\=en-us} lung disease diagnosis dataset contains a total of 1,470 CT scans with four diseases. The annotation process is divided into two stages: two physicians perform the original annotation, then a third independent physician perform the disambiguation to ensure the consistency of the data annotation. There are 3,264 nodules, 3,613 Streak shadows, 4,201 arteriosclerosis or calcification and 1,140 lymph node calcification.

\subsection{Implementation Details}
We transform each 3D lesion to a corresponding 2D view, and resize each view to $ 224 \times 224 $. To generate the different transformed 2D views, we rotate the coordinate systems clockwise with angles of $ 50^{\circ} $, $ 90^{\circ} $, $ 140^{\circ} $ and $ 180^{\circ} $. 
We use the ResNet18 \cite{he2016deep} as the backbone without fully connected layer. Adam \cite{kingma2015adam} with a base learning rate of 0.001 is used to optimize the SCL. The training epoch is set to 1000 with an early-stopping of 50. We set the batch size to 64, and the weight decay rate is $ 1e-4 $. The temperature $ \tau $ is set to 0.07. All the experiments are conducted with PyTorch \footnote{https://pytorch.org/} using a single Nvidia RTX 2080 ti 11GB GPU.

\subsection{Lung Nodule Classification}
We train a simple classifier on features extracted from the frozen pre-trained SCL to evaluate the representations, as shown in Table \ref{table_linear_cls}. The Models Genesis \cite{zhou2019models}, Rubik's Cube+ \cite{zhu2020rubik} and Restoration \cite{chen2019self} are developed for medical images, and the others are developed based on natural images. 

Comparing the existing state-of-the-art SSL methods, the accuracy of the methods based on MIA is higher than the method based on NIA when applying them to CT scans. This proves that MIA is more effective for medical images. For the proposed SCL on LIDC-IDRI, we obtain an accuracy of 89.72\%. Compared to SimCLR (MIA) and Rubik's Cube+, SCL achieves 9\% and 8\% improvement, respectively. Compared to state-of-the-art SSL methods, the highest accuracies are also obtained on both LNDb (82.09\%) and TianChi (90.16\%). The results show that the proposed SCL has superior performance than previous SSL methods.


\begin{figure}[t]
    \centering
    \includegraphics[width=\linewidth]{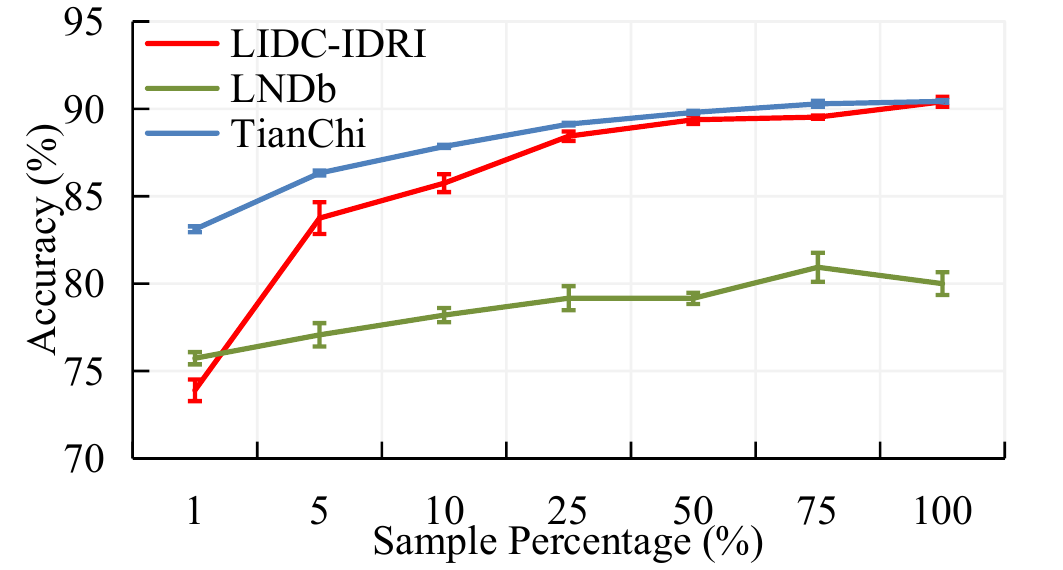}
    \caption{Results of fine-tuning with different numbers of samples.}
    \label{figure_fine-tune}
\end{figure}

\renewcommand{\arraystretch}{1.4}
\begin{table*}[h]
	\centering
	\setlength{\tabcolsep}{14pt}
	\begin{threeparttable}
		\caption{Fine-tuning with 10\% samples on the three datasets ($ Mean(\%) \pm std (\%) $)}
		\label{table_fine-tune}
		\begin{tabular}{l c c c c c c}
			\hline\hline
			 & \multicolumn{2}{c}{LIDC-IDRI} & \multicolumn{2}{c}{LNDb} & \multicolumn{1}{c}{TianChi} \\
			\hline
			 & \textbf{AUC} & \textbf{Accuracy} & \textbf{AUC} & \textbf{Accuracy} & \textbf{Accuracy} \\
			\hline
			Supervised & $ \mathbf{90.90} \pm 0.47 $ & $ 85.03 \pm 0.50 $ & $ 70.65 \pm 0.81 $ & $ 73.44 \pm 0.21 $ & $ 83.34 \pm 0.33 $ \\
			Supervised (3D) & $ 78.88 \pm 0.31 $ & $ 76.60 \pm 0.42 $ & $ \mathbf{77.43} \pm 0.85 $ & $ \mathbf{80.25} \pm 0.53 $ & $ 78.13 \pm 0.75 $ \\
			SimCLR (NIA) \cite{chen2020simple} & $ 85.91 \pm 0.74 $ & $ 84.90 \pm 0.73 $ & $ 73.62 \pm 0.35 $ & $ 72.97 \pm 0.99 $ & $ 79.08 \pm 0.56 $ \\
			SimCLR (MIA) \cite{chen2020simple} & $ 85.47 \pm 0.17 $ & $ 85.46 \pm 0.22 $ & $ 73.85 \pm 0.92 $ & $ 78.34 \pm 0.91 $ & $ 80.72 \pm 0.93 $ \\
			\hline
			\textbf{SCL} (Ours) & $ 90.11 \pm 1.15 $ & $ \textbf{85.75} \pm 1.01 $ & $ 69.16 \pm 1.34 $ & $ 78.20 \pm 0.63 $ & $ \mathbf{87.85} \pm 0.09 $ \\
			\hline\hline
		\end{tabular}
	\end{threeparttable}
\end{table*}

\subsection{Fine-tuning on Three Datasets}
To further evaluate the effectiveness of the SCL, we fine-tune the model with limited annotated samples in each dataset. As shown in Table \ref{table_fine-tune}, we fine-tune the SCL with 10\% datasets, and the accuracies are 85.75\% and 78.20\% on LIDC-IDRI and LNDb, respectively, and they are comparable to Supervised. We also conduct multi-disease diagnosis on TianChi, and the diagnostic accuracy of SCL is higher (4\%) than that of Supervised. Overall, the SCL fine-tuned with 10\% datasets is better than SimCLR and is comparable to the Supervised model.

We also investigate the performance of SCL for fine-tuning with different percentages of samples on the three datasets and summarize the results in Fig. \ref{figure_fine-tune}. According to the overall trend in Fig. \ref{figure_fine-tune}, the accuracy increases with the increase of samples. Even though the samples are very small (e.g., percentage=1\%), the accuracy of SCL also reaches a relatively higher value. It proves that SCL is able to learn high-quality lesion representations, therefore, only a small number of samples are needed to learn the category of lesions. Therefore, we can construct high-performance CAD systems based on SSL with a small number of samples.

\section{Conclusion}
In this paper, we propose the SCL, a novel spiral contrastive learning framework for yielding 3D representations in a computationally efficient manner. Unlike existing 3D SSL methods, SCL is a light-weight network due to we convert each 3D lesion into a 2D view by the information-preserving spiral transformation. Most studies apply the 2D slices in a transverse plane as the input, so it ignores the correlation of information between the adjacent layers. Spiral transformation takes the whole 3D lesion to a 2D plane, which retains the correlation of the features in 3D space. Experimental results on three CT datasets show that the proposed SCL outperforms state-of-the-art SSL methods, indicating the superiority of SCL in \emph{unsupervised representation learning}. Being fine-tuned with a small percentage of the datasets (10\%), our model is comparable to the 3D fully supervised model, demonstrating its superiority in \emph{small-data scenarios} and the potential of reducing the annotation efforts in 3D CAD systems.

The main insufficiency of our work is that SCL is designed for the lesion diagnosis task and is not optimized for the localization task on the whole CT volumes. The effectiveness of SCL methods in lesion localization (often with background dominance) needs to be explored. In the future, we will adapt SCL to more tasks (lesion localization and segmentation) and evaluate it on more imaging modalities such as the magnetic resonance imaging.

\printbibliography

\end{document}